# FOODER: Real-time Facial Authentication and Expression Recognition

*Sabri Mustafa Kahya*, Muhammet Sami Yavuz*, Boran Hamdi Sivrikaya*, Eckehard Steinbach**
*Technical University of Munich, School of Computation, Information and Technology,
Department of Computer Engineering, Chair of Media Technology, Munich Institute of Robotics
and Machine Intelligence (MIRMI)
mustafa.kahya, sami.yavuz, boran.sivrikaya, eckehard.steinbach@tum.de*

## Abstract

Out-of-distribution (OOD) detection is critical for the safe deployment of modern neural network architectures, as it aims to identify samples outside the training domain. In this chapter, we introduce FOODER, a novel real-time, privacy-preserving radar-based framework that combines OOD-based facial authentication with expression recognition. FOODER operates using low-cost frequency-modulated continuous-wave (FMCW) radar and leverages both range-Doppler and micro range-Doppler images to enhance the overall performance. The authentication module of FOODER consists of a multi-encoder multi-decoder Body Part (BP) and Intermediate Linear Encoder-Decoder (ILED) components, enabling accurate classification of a single individual's face as in-distribution (ID) while identifying all other faces as OOD.
Upon successful ID classification, the expression recognition module is activated. FOODER processes the concatenated range-Doppler and micro range-Doppler representations through a ResNet block to determine whether the facial expression is dynamic or static. This categorization is based on the observation that expressions like smiling and shock involve more pronounced and rapid facial movements compared to more subtle or static expressions such as anger and neutral. Accordingly, two specialized MobileViT architectures are employed: one for classifying dynamic expressions (smile, shock) and the other for classifying static expressions (neutral, anger). This hierarchical design enables both robust facial authentication and fine-grained expression recognition, while preserving user privacy by relying solely on radar data. On our dataset collected with 60 GHz short-range FMCW radar, FOODER achieves an Area Under the Receiver Operating Characteristic (AUROC) curve of 94.13% and a False Positive Rate of 18.12% at a True Positive Rate of 95% (FPR95) for the authentication task, and an average classification accuracy of 94.70% for facial expression recognition. Additionally, FOODER outperforms state-of-the-art OOD detection methods in terms of facial authentication, achieves better facial expression recognition accuracy than several transformer-based architectures, and operates efficiently in real-time.
*Index Terms*—Facial authentication, facial expression recognition out-of-distribution detection, 60 GHz FMCW radar, deep neural networks

# 1. INTRODUCTION

Facial Expression Recognition (FER) and facial authentication are two fundamental tasks within human-centered computing, enabling systems to interpret emotional states and verify individual identities, respectively. FER focuses on identifying emotional cues from facial expressions [1], playing a key role in enhancing Human-Computer Interaction (HCI) [2], virtual reality experiences [3], and digital entertainment applications [4], [5]. Meanwhile, facial authentication systems aim to verify user identity, supporting applications in security, device access, and digital identity management.

As a critical and popular computer vision application, facial authentication is now widely integrated into everyday technologies. For instance, many smartphones employ facial authentication via front-facing cameras to unlock screens, ensuring only authorized users can access the device. Numerous research efforts [6], [7], [8] and practical software solutions have utilized RGB image-based sensors for this purpose. Although these systems have achieved impressive performance, they inherently rely on visual data, raising significant privacy concerns. Despite the success of vision-based methods in both FER [9], [10], [11], and facial authentication, they remain vulnerable to environmental factors such as lighting variations, occlusion, and pose changes. Furthermore, capturing and storing visual information raises concerns around user privacy and data security, especially as computer vision technologies become more pervasive across industries like autonomous driving, healthcare, and retail.

Radar sensors have gained increasing popularity due to their resilience under challenging environmental conditions such as fog, smoke, and poor lighting, and their inherent ability to preserve user privacy. These properties make radar an attractive sensing modality for a wide range of applications, including human presence detection, people counting, gesture recognition, and even vital sign monitoring [12], [13], [14], [15].

Short-range radar systems, in particular, are becoming indispensable tools in both academic research and industrial solutions, especially for indoor environments.
Facial authentication, a binary classification task, naturally fits within the framework of One-vs-All (OvA) problems. However, given the practically infinite number of "all" class members—i.e., individuals who are not the target identity—facial authentication inherently becomes an out-of-distribution (OOD) detection problem. Traditional classification models, designed to recognize only a closed set of known classes, struggle with unknown identities. OOD detection reframes the facial authentication challenge as a binary classification task: distinguishing in-distribution (ID) samples, belonging to the enrolled user, from OOD samples, representing all other individuals.
OOD detection [16], [17], [18], [19], [20] is critical for safely deploying machine learning models, especially in safety-critical applications such as autonomous driving, healthcare, robotics, and biometric authentication. Without effective OOD handling, neural networks are prone to making overconfident and potentially catastrophic predictions on unfamiliar inputs. A detector assigns a confidence score to each sample in a typical OOD detection setting. A threshold is determined based on a validation set containing both ID and OOD examples (e.g., ensuring 95% true positive rate for ID samples). During testing, if a sample's score exceeds the threshold, it is classified as ID; otherwise, it is rejected as OOD.

This work proposes a radar-based framework that unifies facial authentication and facial expression recognition within a single, privacy-preserving system. By leveraging low-cost frequency-modulated continuous-wave (FMCW) radar and advanced deep learning architectures, we ensure both secure identity verification and rich emotional understanding without relying on visual imagery. Our system uses radar range-Doppler images (RDIs) and micro range-Doppler images (micro-RDIs) as inputs.

For facial authentication, we adopt a reconstruction-based OOD detection strategy. Specifically, we utilize a multi-encoder, multi-decoder architecture. The system assigns reconstruction error-based scores to each sample to determine whether it belongs to the ID class (scores below a predefined threshold) or to the OOD class (scores above the threshold).

The facial expression recognition module is activated if the detected sample belongs to the ID class. First, RDIs and micro-RDIs are concatenated and passed through a ResNet block to classify the expression as dynamic or static. Based on this classification, two specialized MobileViT networks are employed: one trained to distinguish dynamic expressions (smile and shock) and the other trained for static expressions (anger and neutral). This two-stage pipeline ensures accurate recognition of the user's emotional state while maintaining real-time performance and preserving user privacy.

This work is built upon our previous contribution, RFOOD [29], a real-time facial authentication system. Building on RFOOD's foundation, we introduce the following new contributions:

- We extend RFOOD by integrating a facial expression recognition module on top of the authentication pipeline, enabling a unified framework that supports both user authentication and expression understanding based solely on radar data.
- We design a novel hierarchical FER architecture that integrates facial expression recognition by employing a ResNet-based classifier to distinguish between dynamic (smile, shock) and static (neutral, anger) expressions, followed by two specialized MobileViT models dedicated to classifying specific expressions within each category. This design creates a unified framework for secure and emotionally-aware human–machine interaction.
- We construct a new, diverse radar-based dataset for training and evaluating the FER module.

- Overall, FOODER achieves an AUROC of 94.13% for OOD detection and an average expression recognition accuracy of 94.70% on our radar-based dataset, demonstrating strong performance across both authentication and expression recognition tasks—while operating in real time.

## 2. RELATED WORK

Several studies have explored radar-based face authentication, each contributing unique approaches. For example, [21] utilized a 61 GHz millimeter-wave radar in combination with a deep neural network (DNN) to classify faces. Their dataset, created from eight individuals positioned at various distances and angles relative to the radar, was formed by collecting signals from multiple antenna elements. This method achieved a classification accuracy of 92.0% by merging signals from each receiver. In another study, [22] developed a facial authentication system using Convolutional Neural Networks (CNNs). They collected data from three individuals, both with and without cotton masks, positioned 30 cm from a 61 GHz FMCW radar. Their findings include results for both masked and unmasked conditions. Additionally, [23] introduced a system with 32 transmit and 32 receive antennas, employing a dense autoencoder architecture for personalized facial authentication through a one-class classifier, trained on 200 faces. [24] used the same dataset but improved the model by combining a convolutional autoencoder with a random forest classifier. In a separate approach, [25] proposed a one-shot learning-based Siamese network for facial recognition using a 61 GHz FMCW radar. Their study, which involved eight individuals, achieved a high classification accuracy of 97.6%. [26] employed four cascaded 77 GHz mmWave radars, each with 3 transmit and 4 receive antennas, to create a 12 Tx 16 Rx antenna system. Using data from nine individuals positioned 60 cm to 80 cm from the radar, they extracted sparse point clouds for facial recognition, achieving an outstanding 98.69% classification accuracy by training a PointNet-based model.

Kahya et al. [27] introduced FOOD, a short-range 60 GHz FMCW radar-based facial authentication system that jointly performs multi-class in-distribution (ID) face classification and binary OOD detection. The system employs a convolutional one-encoder, multi-decoder architecture that processes raw ADC radar data. It was trained solely on data from three male ID subjects and evaluated on both ID and 13 OOD individuals (10 male, three female), achieving promising results across both tasks.

FARE [28] framework introduces a unified radar-based solution that simultaneously addresses face recognition and OOD detection without compromising ID classification accuracy. Data were collected from 16 individuals using a 60 GHz FMCW radar: five were treated as ID (four males, one female), and the remaining ten (nine males, two females) as OOD. FARE consists of two main architectural components. Primary Path (PP) includes modality-specific feature extractors for both RDIs and micro-RDIs. Intermediate features are fused via a combined extractor block to generate robust embeddings. The PP is trained with a triplet loss using the Adamax optimizer to ensure discriminative separation among ID faces. Intermediate Paths (IPs) are linear autoencoders that are appended to form a reconstruction-based mechanism for OOD detection at the end of each layer of the PP. These paths are trained using only ID samples with Mean Absolute Error (MAE) loss. During inference, reconstruction errors from IPs are compared against a threshold to identify OOD samples. The system employs a two-stage training strategy: first, the PP is optimized for face classification using triplet loss; next, with the PP frozen, the IPs are independently trained for OOD detection. Final classification of ID faces is performed via a K-Nearest Neighbors (KNN) algorithm based on learned embeddings.

RFOOD [29] is a privacy-preserving, real-time OOD detection framework for facial authentication using low-cost FMCW radar. It utilizes both RDIs and micro-RDIs to improve robustness. The architecture features a multi-encoder, multi-decoder backbone (BP) and two Intermediate Linear Encoder-Decoder (ILED) modules. Each ILED compresses intermediate feature maps into a latent space and reconstructs them to enhance overall detection performance. The system is trained using only ID samples and distinguishes a single in-distribution identity from unseen individuals as OOD. RFOOD outperformed state-of-the-art OOD methods on standard metrics while operating in real time. FOODER is built on RFOOD.

Short-range radar has also been explored for facial expression recognition. [30] proposed a system that uses raw mmWave signals to detect fine facial muscle movements, enabling recognition of seven common expressions through a dual-localization mechanism that isolates the face and filters noise. Similarly, [31] employed 3D range-angle spectra from mmWave radar with 3 Tx and 4 Rx antennas to reconstruct facial geometry and track expression changes using triplet loss embeddings. FERT [32] presents a short-range FMCW radar-based system that recognizes smile, anger, neutral, and no-face expressions in real-time and across individuals. It simultaneously leverages RDIs, micro-RDIs, Range-Azimuth Images (RAIs), and Range-Elevation Images (REIs), each processed through dedicated feature extractor blocks. Intermediate features are fused—RDIs with micro-RDIs, and RAIs with REIs—then passed through modality-specific intermediate extractors and finally integrated via a ResNet block for classification. FERT achieves an average accuracy of 98.91% on data captured with a 60 GHz radar.

OOD detection was first introduced in a groundbreaking study [33], which utilized the maximum softmax probabilities to distinguish between ID and OOD samples. The authors observed that ID samples typically have higher softmax scores. Building on this, ODIN [34] enhanced the detection process by incorporating input perturbations and temperature scaling, which improved the softmax scores of ID samples. MAHA [35] leverages the intermediate feature representations within a model for OOD detection and introduces a distance-based approach to identify OOD instances. Similarly, FSSD [36] relies on intermediate outputs and claimed that OOD samples behave more like uniform noise compared to ID ones. [37] proposed a straightforward method using K-nearest neighbors (KNN), utilizing distance information from the final dense layer. The energy-based OOD detector [38] uses energy scores calculated through the *logsumexp* function, claiming that ID samples exhibit lower energy scores than OOD samples. GradNorm [39] distinguishes ID from OOD instances by analyzing gradient magnitudes, which are derived from the backpropagation of the Kullback-Leibler (KL) divergence between the softmax output and a uniform distribution. ReAct [40] introduced a simple activation truncation technique that can be applied alongside any OOD detector to enhance its detection performance. In another study [41], two powerful OOD detection methods were introduced: one based on maximum logit values and the other utilizing minimum KL divergence. Some works benefit from a small set of OOD samples during training to improve the separation between ID and OOD samples [42], [43].

While the majority of SOTA OOD detection methods focus on image data, a few studies have explored this problem in the context of radar data. In [44], an OOD detection model was proposed using RDIs as input to a patch-based autoencoder architecture, aiming to distinguish between a walking person and other types of moving objects. Building on this, MCROOD [45] introduced a one-encoder, multi-decoder framework that extends the previous work by also considering sitting and standing individuals as ID cases. Separately, [12] addressed both human presence detection and OOD detection simultaneously, without relying on specific human activity types. Furthermore, [46] focused on identifying ID human activities while accurately detecting OOD instances.

## 3. RADAR SYSTEM DESIGN

This study employs Infineon's BGT60TR13C, a 60 GHz FMCW radar chipset operating between 57 GHz and 64 GHz with configurable chirp duration ($T_c$). The module features one transmit (Tx) antenna and three receive (Rx) antennas arranged in an L-shaped formation, with half-wavelength spacing between adjacent antennas. The Tx antenna has a gain of 10 dBi, while each Rx antenna provides 6 dBi gain.

The radar operates based on FMCW principles. A voltage-controlled oscillator (VCO), stabilized by a phase-locked loop (PLL) using an 80 MHz reference signal, produces linear frequency chirps ($N_c$) from 57 GHz to 64 GHz. Frequency modulation is achieved by adjusting a divider and applying a tuning voltage between 1V and 4.5V. The transmitted chirp signal is expressed as

$$s(t) = \exp\left(j2\pi\left(f_c t + \frac{S}{2}t^2\right)\right), \quad \forall 0 < t < T_c$$

where $f_c$ denotes the center frequency, and $S$ is the chirp rate, calculated as $S = \frac{B}{T_c}$, with $B$ being the bandwidth. When this signal reflects off objects within the radar's field of view, it experiences a time delay proportional to the object's range and motion. The delayed reflections are received by the antennas. The received signal is mixed with the transmitted chirp and passed through a low-pass filter to isolate the intermediate frequency (IF) signal. This IF signal is then digitized using a 12-bit Analog-to-Digital Converter (ADC) operating at a sampling rate of 2 MHz. The digitized data is structured into frames of size $N_c \times N_s$, where each column represents $N_c$ slow-time samples, and each row contains $N_s$ fast-time samples. Considering the three receive antennas, the resulting data forms a three-dimensional matrix with dimensions $N_{Rx} \times N_c \times N_s$, which is then used for subsequent signal processing. The detailed configuration of the FMCW radar utilized in this work is provided in Table 1. Based on this setup, the radar achieves a range resolution of $\delta r = \frac{c}{2B} = 3.75$ cm, where $c \approx 3 \times 10^8$ m/s is the speed of light. The maximum unambiguous range is given by $R_{max} = \frac{N_s}{2} \times \delta r = 2.4$ m. The maximum measurable velocity is $v_{max} = \frac{c}{2 f_c T_{cc}} \approx 6.38$ m/s, and the velocity resolution is $\delta v = \frac{c}{2 f_c \left(\frac{N_c}{2}\right) T_{cc}} \approx 0.20$ m/s.

Table 1: FMCW Radar Configuration Parameters

| Configuration name | Symbol | Value |
|---|---|---|
| Number of transmit antennas | $N_{Tx}$ | 1 |
| Number of receive antennas | $N_{Rx}$ | 3 |
| Chirps per frame | $N_c$ | 64 |
| Samples per chirp | $N_s$ | 128 |
| Frame period | $T_f$ | 50 ms |
| Chirp to chirp time | $T_{cc}$ | 391.55 µs |
| Ramp start frequency | $f_{min}$ | 58.0 GHz |
| Ramp stop frequency | $f_{max}$ | 62.0 GHz |
| Bandwidth | $B$ | 4 GHz |

### 3.1. Pre-processing

This study utilizes RDIs and micro-RDIs as inputs to the proposed architecture, both generated from the digitized radar signal of size $\mathbf{N_{Rx} \times N_c \times N_s}$. To generate the RDI, a Range-FFT is first applied along the fast-time axis to extract range information. Mean removal across the three Rx channels yields a single-channel representation. Next, frame-wise Moving Target Identification (MTI) is employed to eliminate static reflections. A Doppler-FFT is then performed along the slow-time dimension to extract motion-induced phase variations, resulting in the RDI. The micro-RDI follows a similar initial step with a Range-FFT for extracting range information. To reduce noise, eight range spectrograms are stacked, and mean subtraction is applied along both the fast-time and slow-time dimensions. A Sinc filter is then used to enhance target signal components. Finally, a Doppler-FFT along the slow-time axis produces the micro-RDI. As a final preprocessing step, both RDIs and micro-RDIs are refined using E-RESPD [12] which enhances the representations to more effectively capture facial movements.

# 4. PROBLEM STATEMENT AND METHOD

The proposed system integrates personalized facial authentication and facial expression recognition pipelines. The facial expression recognition module is built on top of the existing facial authentication framework, RFOOD. The overall system performs two primary tasks: a binary classification for facial authentication, distinguishing between the ID face (the known individual) and OOD faces (i.e., all other individuals), and a multi-class facial expression recognition task that classifies the facial expression of the authenticated person as smile, shock, anger, or neutral.

## 4.1. Architecture and Training

**RFOOD:**
The central body part (BP) of the RFOOD architecture employs a multi-encoder multi-decoder design, specifically featuring two parallel branches with identical structures. Each encoder within these branches progressively reduces the spatial dimensions of the input through a sequence of three convolutional layers, utilizing 16, 32, and 64 filters respectively, along with 3x3 kernels, a padding of 1, and a stride of 2 for downsampling. Each convolutional layer is followed by batch normalization and a LeakyReLU activation function. The decoders replicate this structure but employ transposed convolutional layers to reconstruct the input. They use filters of 64, 32, and 16 with the same kernel size and padding, a stride of 2, and output padding of 1 for upsampling. Like the encoders, each transposed convolutional layer is followed by batch normalization and a LeakyReLU activation, except for the final layer, which incorporates a sigmoid activation to generate the output. The RFOOD architecture also includes auxiliary intermediate linear encoder-decoder units (ILEDs).

Each ILED module sits right before the last upsampling step in its corresponding decoder. Within this module, the intermediate feature maps are first flattened into a single long vector. This vector then goes through a dense layer that compresses it into a smaller, 128-dimensional latent representation, with batch normalization applied for better training. A second fully connected layer then reconstructs the vector back to its flattened shape, followed once again by batch normalization. The final step reshapes this output back into its initial spatial configuration. By adding this brief compression and decompression stage, the network gains more expressive intermediate feature representations. Please see the upper part of Figure 1 for the high-level overview of the facial authentication module of FOODER.

RFOOD processes two input modalities (RDIs and micro-RDIs). Within BP, one encoder-decoder pair is responsible for handling RDIs, while the other is dedicated to processing micro-RDIs. For each of these encoder-decoder pairs, a separate mean squared error (MSE) loss is computed, yielding two individual loss components. The total loss for the BP is expressed as:

$$\mathcal{L}_{BP} = \frac{1}{n}\sum_{j\in\{R,mR\}} \sum_{i=1}^{n}\left(\mathbf{X}_j^{(i)} - D_j\left(E_j\left(\mathbf{X}_j^{(i)}\right)\right)\right)^2,$$

where $\mathbf{X}_j^{(i)}$ is the input, $j$ is the index for RDI ($R$) and micro- RDI ($mR$), $n$ is the batch size, $E_j$ and $D_j$ demonstrate the encoders and decoders of BP.

The ILED modules are designed to process intermediate feature maps by first compressing them into a compact latent representation and then reconstructing them back to their original dimensionality. One ILED is dedicated to the RDI branch, while the other handles micro-RDI data. By introducing this additional encoding-decoding stage within the decoder, the network benefits from enhanced representational capacity, ultimately improving its detection performance. Each ILED introduces its own MSE-based reconstruction loss—one for RDIs and another for micro-RDIs. The combined ILED loss is defined as:

$$\mathcal{L}_{ILED} = \frac{1}{n}\sum_{k\in\{il_R,il_{mR}\}} \sum_{i=1}^{n}\left(\mathbf{F}_k^{(i)} - D_k\left(E_k\left(\mathbf{F}_k^{(i)}\right)\right)\right)^2,$$

where $\mathbf{F}_k^{(i)}$ is the input intermediate feature representation, $k$ is the index for ILEDs of RDIs ($il_R$) and micro-RDIs ($il_{mR}$). $n$ is the batch size. $E_k$ and $D_k$ are the linear encoder-decoder pairs for their respective type.

The overall loss function of the network comprises four mean squared error (MSE) components: two derived from the BP—one corresponding to RDIs and the other to micro-RDIs—and two from the ILEDs, which account for the reconstruction of intermediate features from both modalities. The total loss is defined as:

$$\mathcal{L} = \mathcal{L}_{BP} + \mathcal{L}_{ILED}$$

The entire RFOOD architecture is trained exclusively on ID samples. In line with the principles of OOD detection, the model is never exposed to any OOD data during training. Optimization of all four MSE components is performed jointly using the Adamax optimizer.

### Expression Recognition:

As stated, the facial expression recognition pipeline is built on top of the RFOOD framework. In this design, the RFOOD architecture remains frozen, and only the expression recognition pipeline is trained. The proposed system features a novel architecture that leverages the strengths of both convolutional neural networks and transformer-based architectures. Specifically, it incorporates a standard ResNet18 block, along with two specialized transformer-based architectures—MobileViT and MobileViT-v2. The input modalities remain consistent with those used in RFOOD, namely RDIs and micro-RDIs.

The architecture is composed of three separate classification blocks, each targeting a distinct binary classification task:

**Dynamic-Static Facial Movement Classification (ResNet Block):**
This block employs the ResNet18 architecture to differentiate between dynamic and static facial expressions. The dynamic category includes expressions such as smile and shock, which involve significant facial movement, while the static category includes anger and neutral expressions, characterized by minimal motion. RDI and micro-RDI modalities are first concatenated and then fed into the ResNet18 block. The network is trained using a binary cross-entropy loss function with the Adam optimizer.

**Smile-Shock Expression Classification (MobileViT-v2 Block):**
This block focuses exclusively on differentiating between smile and shock expressions, both of which are classified as dynamic. It utilizes the MobileViT-v2 architecture, again using concatenated RDI and micro-RDI inputs. The model is trained separately for this binary classification task using cross-entropy loss and the Adam optimizer.

**Anger-Neutral Expression Classification (MobileViT Block):**
This block is responsible for distinguishing between anger and neutral expressions, which fall under the static expression category. It utilizes the MobileViT architecture, also operating on merged RDI and micro-RDI inputs. Like the other blocks, it is trained for binary classification using cross-entropy loss and the Adam optimizer.

Each block is trained independently to optimize for its specific classification task. The ResNet18 block is trained on the entire training dataset, encompassing all four expression types (smile, shock, anger, neutral), to classify expressions as dynamic or static. In contrast, the MobileViT and MobileViT-v2 blocks are trained on subsets of the data. The MobileViT-v2 block uses only smile and shock samples, while the MobileViT block uses anger and neutral samples. This division enables the model to specialize more effectively in distinguishing between closely related expressions, leading to improved overall classification performance. Please see the lower part of Figure 1 for the high-level overview of the facial expression recognition module of FOODER.

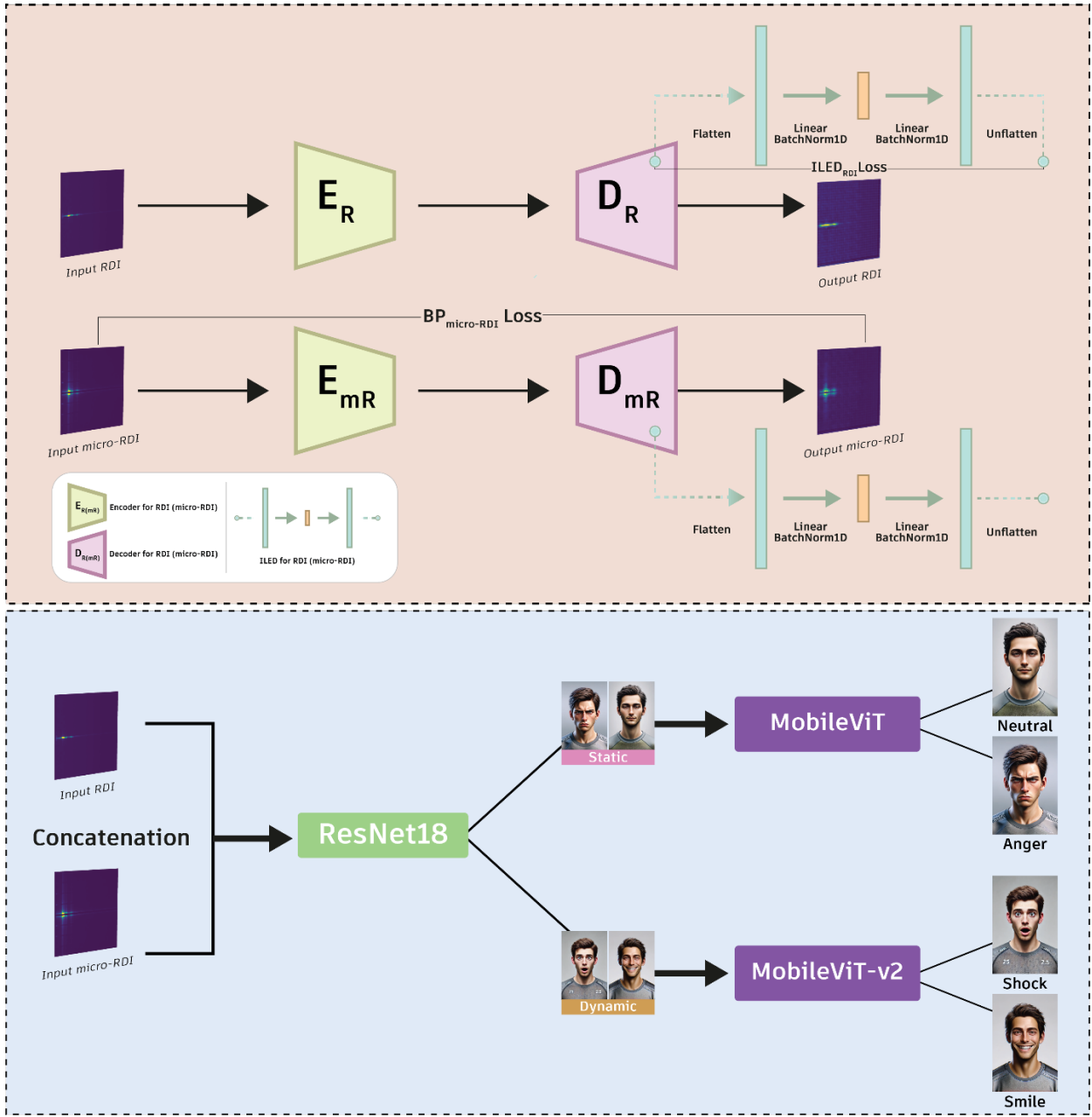

Figure 1. This figure provides a high-level overview of the FOODER pipeline. The upper part of the figure represents the facial authentication module (RFOOD), while the lower part illustrates the facial expression recognition module. The RFOOD module comprises the Body Part (BP) and Intermediate Linear Encoder-Decoders (ILEDs). The BP consists of two encoder-decoder pairs specialized for processing Range-Doppler Images (RDIs) and micro-RDIs: the upper pair handles RDIs, while the lower pair is dedicated to micro-RDIs. Each ILED component is positioned right before the final layer of each decoder in the BP. These components are responsible for encoding and decoding intermediate feature representations. Specifically, the upper ILED processes intermediate features of RDIs, while the lower ILED handles those of micro-RDIs. Overall, the RFOOD module functions as a robust OOD detector and facial authenticator. The facial expression recognition module (FER) consists of three distinct blocks: ResNet18, MobileViT, and MobileViT-v2. Like the RFOOD module, all three blocks utilize both RDIs and micro-RDIs. The ResNet18 block determines whether the facial expression is dynamic or static. If the expression is identified as dynamic, the MobileViT-v2 block is triggered to classify the expression as either smile or shock. If the expression is static, the MobileViT block is activated to classify it as either anger or neutral. Importantly, the FER module is only activated if the RFOOD module successfully authenticates the individual as ID.

## 4.2. OOD Detection and Expression Classification

In our experimental setup, the ID category comprises frames containing the face of a single individual, whereas the OOD category includes facial data from other individuals. For OOD detection, we utilize the total reconstruction MSE as a scoring metric. Given that the network is trained solely on ID data, it is expected to produce lower reconstruction errors for ID samples compared to OOD samples. To support real-time inference, a fixed threshold is manually selected such that 95.0% of the ID samples are correctly identified.

During inference, the total reconstruction loss of an incoming test sample is computed. If this value falls below the predefined threshold, the sample is classified as ID; otherwise, it is considered OOD. The detailed procedure is also illustrated in Algorithm 1.

**Algorithm 1:** RFOOD Pseudocode

**if**
$$MSE(X_R, D_R(E_R(X_R))) \\ + MSE(X_{mR}, D_{mR}(E_{mR}(X_{mR}))) \\ + MSE(F_{il\_R}, D_{il\_R}(E_{il\_R}(F_{il\_R}))) \\ + MSE(F_{il\_mR}, D_{il\_mR}(E_{il\_mR}(F_{il\_mR}))) > \text{threshold}$$
  **then**
    $X \leftarrow$ OOD
  **else**
    $X \leftarrow$ ID
  **end if**

Although the facial expression classification module incorporates both convolutional and transformer-based architectures, the overall pipeline maintains real-time performance due to the lightweight nature of the ResNet and MobileViT blocks. The expression recognition process is initiated only after the facial authentication module successfully verifies the individual as an ID user.

Once authentication is confirmed, the input modalities are concatenated and passed to the ResNet18 block. This block determines whether the facial movement corresponds to a dynamic (e.g., smile or shock) or static (e.g., anger or neutral) expression. If the expression is classified as dynamic, the specialized MobileViT-v2 architecture is triggered to distinguish between smile and shock. Conversely, if the ResNet18 block identifies the expression as static, the MobileViT block is activated to determine whether the expression corresponds to anger or neutral. Each of these blocks uses the same merged input modalities for inference.

Through this sequential decision-making process, the system accurately classifies the current facial expression of the authenticated user as either smile, shock, anger, or neutral, all while preserving real-time operability.

## 5. EXPERIMENTS

In this study, we address two core tasks using FMCW radar data: facial authentication and facial expression recognition. To support these objectives, we collected a custom RF dataset focused on the facial region using the BGT60TR13C 60 GHz FMCW radar sensor. During data collection, the radar was positioned 25 cm from each subject's face, and participants rested their chins on a table to ensure consistent alignment. Recording sessions lasted between 1 and 5 minutes and were conducted in different rooms and at various times of day to introduce environmental diversity. No facial accessories, such as eyeglasses, were worn during the sessions.

The dataset includes both identity data (ID vs. OOD faces) and four distinct facial expressions. For the authentication task, ID data consists of recordings from a single male participant, while OOD data includes recordings from six other male individuals.

For expression recognition, we focus exclusively on the ID subject. We recorded four facial expressions: smile, shock, anger, and neutral. The smile expression featured raised cheeks and visible teeth; shock expression involved wide eyes and an open mouth; anger was portrayed through clenched teeth, furrowed brows, and narrowed eyes; and the neutral expression was characterized by the absence of deliberate facial movement. Each expression was performed multiple times per session to ensure consistency.

For facial authentication, we used 71116 ID frames for training, and a test set comprising 26501 ID frames and 99743 OOD frames. For the facial expression recognition task, the training set includes 102174 frames of smiling, 84082 frames of shock, 90639 frames of anger, and 99645 frames of neutral expressions. The test set comprises 69464 smile, 41270 shock, 55228 anger, and 40736 neutral frames. All participants provided written consent, and the dataset will be made publicly available [https://syncandshare.lrz.de/getlink/fiVXWNxoYss8rRk2kZTUws].

We approach the facial authentication task as an OOD detection problem. In this setup, the ID class consists of a single enrolled subject, while all other subjects are treated as OOD. Accordingly, we evaluate authentication performance using standard OOD detection metrics. Specifically, we report AUROC, which measures the area under the receiver operating characteristic curve; $AUPR_{IN/OUT}$, which refers to the area under the precision-recall curve when either ID or OOD samples are treated as the positive class; and FPR95, which is the false positive rate (FPR) at a true positive rate (TPR) of 95%. We also include Test Time, defined as the total inference time in seconds required to evaluate all test samples. For the facial expression recognition task, we report the average classification accuracy, along with a confusion matrix that provides detailed insight into per-class performance.

We conduct separate experiments for both the facial authentication and facial expression recognition components of the proposed system.

Given that our robust facial authentication framework, RFOOD, also functions as an OOD detector, we compare its performance against several state-of-the-art (SOTA) OOD detection methods. Most existing SOTA techniques are designed to operate on standard classifiers such as ResNet. Therefore, for a fair comparison, we train a ResNet34 model using a one-class classification approach, utilizing the same ID data used for training RFOOD, without exposing the model to any OOD samples during training. We then apply nine different OOD detection techniques on top of the trained ResNet34 for evaluation.

To ensure fairness, we use the same ID (training and testing) and OOD (testing) datasets employed in the RFOOD pipeline across all methods. As shown in Table 2, RFOOD (the facial authentication module of FOODER) outperforms existing SOTA OOD detection approaches in terms of standard OOD detection metrics as well as inference time.

Table 2: Performance comparison with SOTA methods. All values are shown in percentages. ↑ indicates that higher values are better, while ↓ indicates that lower values are better.

| Method | $AUROC$ ↑ | $AUPR_{IN}$ ↑ | $AUPR_{OUT}$ ↑ | $FPR95$ ↑ | $Test\ Time$ ↓ |
|---|---|---|---|---|---|
| MSP [33] | 49.73 | 36.13 | 63.52 | 94.59 | 63 |
| ODIN [34] | 49.82 | 36.34 | 63.78 | 94.83 | 243 |
| ENERGY [38] | 27.62 | 26.03 | 49.06 | 100.0 | 63 |
| MAHA [35] | 85.15 | 79.48 | 90.20 | 61.99 | 806 |
| FSSD [36] | 61.92 | 39.67 | 79.18 | 69.06 | 948 |
| GRADNORM [39] | 49.19 | 35.99 | 64.72 | 95.75 | 184 |
| REACT [40] | 28.19 | 26.47 | 49.19 | 100.0 | 64 |
| MAXLOGIT [41] | 44.26 | 31.29 | 61.64 | 95.09 | 64 |
| KL [41] | 50.16 | 36.62 | 63.78 | 95.02 | 63 |
| **RFOOD** | **94.13** | **85.83** | **97.02** | **18.12** | **7** |

In addition to offline experiments, we evaluate the RFOOD pipeline in real-time settings, including tests with individuals not present in the original dataset and with subjects wearing realistic face masks in attempts to deceive the system. A real-time demonstration video showcasing only the RFOOD component of our study is available here [https://youtu.be/PI6bkqvjn28].

We also performed a set of ablation studies for the facial authentication section. The first study evaluates the positive impact of using both input modalities—RDIs and micro-RDIs—together during training and inference, as opposed to using each modality separately. The second study investigates the effectiveness of jointly using the ILEDs and BP, rather than using either component alone.

Table 3 presents the results for three configurations: using only RDIs, using only micro-RDIs, and using both modalities together. As clearly demonstrated, the combined usage of both input modalities yields better performance than using either modality alone. Here, separate usage refers to employing only a single encoder-decoder pair, along with its corresponding ILED module, tailored for one specific input modality. For instance, using only the encoder-decoder and ILED specialized for the RDI input.

Table 3: Ablation study to show the impact of simultaneous usage of RDIs and micro-RDIs

| Method | $AUROC \uparrow$ | $AUPR_{IN} \uparrow$ | $AUPR_{OUT} \uparrow$ | $FPR95 \uparrow$ |
|---|---|---|---|---|
| RDI | 85.74 | 76.40 | 91.81 | 47.37 |
| micro-RDI | 90.42 | 81.18 | 94.84 | 30.15 |
| **RDI+micro-RDI** | **94.13** | **85.83** | **97.02** | **18.12** |

Table 4, on the other hand, illustrates the benefit of combining both the BP and ILED components for OOD detection. The results clearly indicate that the joint usage of BP and ILED modules significantly improves performance compared to using either component in isolation.

For the RFOOD part, our ablation studies confirm that the best performance is achieved when both input modalities (RDIs and micro-RDIs) are used in conjunction with the full network architecture, including both the BP and ILED modules.

Table 4: Ablation study to show the impact of the simultaneous usage of BP and ILEDs

| Method | $AUROC \uparrow$ | $AUPR_{IN} \uparrow$ | $AUPR_{OUT} \uparrow$ | $FPR95 \uparrow$ |
|---|---|---|---|---|
| BP | 90.52 | 78.94 | 94.80 | 32.81 |
| ILEDs | 92.11 | **86.31** | 95.70 | 27.23 |
| **BP+ILEDs** | **94.13** | 85.83 | **97.02** | **18.12** |

For the FER module of FOODER, we perform a multi-class classification task using a set of binary classification stages across distinct facial expressions. Accordingly, we utilize the accuracy metric to evaluate performance. The confusion matrix shown in Figure 2 provides a detailed overview of the pipeline's classification capabilities. Also, the average dynamic-static classification accuracy from the ResNet18 block is 98.93%.

Given our unique architecture, comprising a convolutional classifier module and transformer-based components, we demonstrate the benefits of our modular approach. Specifically, our pipeline performs binary classification at each stage, which yields better results than using a single classifier for multi-class classification.

To further support our claim that decomposing the task into binary classifications enhances performance, we compare our modular approach with well-known transformer architectures trained in a multi-class classification manner. For fairness, the transformer models are trained using the same training dataset as our FER module and evaluated on the same test samples.

Table 5 summarizes the results of this comparison. Our architecture not only achieves higher accuracy, as shown in the table, but also maintains real-time performance. Notably, some of the competing models listed in the table are computationally heavy and unsuitable for real-time deployment, yet still underperform relative to our proposed pipeline. A real-time demonstration video of our entire FOODER pipeline, including both RFOOD and FER components, is available here [https://youtu.be/B_jXGAz0nQ8]

Table 5: Performance Comparison with Popular Transformer Architectures on FER module of FOODER

| Model | Smile ↑ | Shock ↑ | Anger ↑ | Neutral ↑ | Average ↑ |
|---|---|---|---|---|---|
| Twins [47] | 84.18 | 94.65 | 92.57 | 86.60 | 88.99 |
| MobileViT [48] | 86.04 | 92.79 | 91.69 | 86.81 | 89.05 |
| MobileViT-v2 [49] | 79.47 | 92.12 | 93.07 | 79.82 | 85.70 |
| DeiT [50] | 82.93 | 93.64 | **96.49** | 87.42 | 89.57 |
| Swin [51] | 86.40 | 92.53 | 89.99 | 90.89 | 89.47 |
| CaiT [52] | 22.40 | 63.78 | 82.50 | 50.14 | 52.19 |
| XCiT [53] | 84.32 | 95.85 | 92.86 | 87.40 | 89.51 |
| **Ours** | **94.06** | **96.11** | 94.42 | **94.76** | **94.70** |

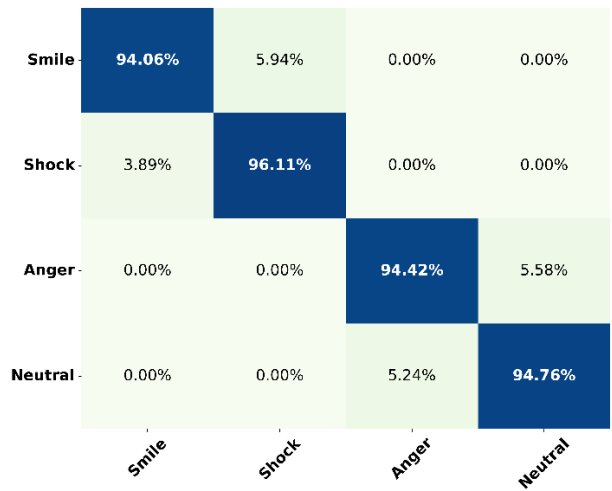

*Figure 2. Confusion matrix to demonstrate the classification performance of FER module of FOODER.*

## 6. CONCLUSION

In this chapter, we introduced FOODER, a novel real-time radar-based framework that unifies facial authentication and expression recognition while preserving user privacy. By utilizing low-cost 60 GHz FMCW radar and exploiting both range-Doppler and micro range-Doppler representations, our method effectively distinguishes in-distribution from out-of-distribution (OOD) faces and classifies facial expressions with high accuracy. The framework's modular architectural components enable hierarchical decision-making optimized for both OOD detection and expression recognition. Experimental results demonstrate that FOODER not only delivers competitive performance in OOD detection and expression recognition but also outperforms state-of-the-art methods on standard benchmarks. With its privacy-preserving design and real-time capability, FOODER sets a new direction for secure and efficient radar-based human-computer interaction systems.